\documentclass[10pt,twocolumn,letterpaper]{article}

\usepackage[pagenumbers]{cvpr} 

\definecolor{cvprblue}{rgb}{0.21,0.49,0.74}
\usepackage[utf8]{inputenc} 
\usepackage[T1]{fontenc}    
\usepackage{url}            
\usepackage{booktabs}       
\usepackage{amsfonts}       
\usepackage{nicefrac}       
\usepackage{microtype}      
\usepackage{xcolor}         
\usepackage{wrapfig} 

\usepackage{amsmath}

\usepackage{multirow}
\usepackage{array}
\usepackage{graphicx} 
\usepackage{caption}
\usepackage{amssymb}
\usepackage{float}
\usepackage{algorithm}
\usepackage{algpseudocode}
\usepackage{adjustbox}
\usepackage{pifont}
\usepackage[pagebackref,breaklinks,colorlinks,allcolors=cvprblue,urlcolor=magenta]{hyperref}
\usepackage[accsupp]{axessibility}

\def\confName{CVPR}
\def\confYear{2026}

\title{Environmental Understanding Vision-Language Model for Embodied Agent}
\author{Jinsik Bang \;\;\; Jaeyeon Bae \;\;\; Donggyu Lee \;\;\; Siyeol Jung \;\;\; Taehwan Kim\\
UNIST\\
{\tt\small \{bang, qowodussla, leedongkyu2019, siyeol, taehwankim\}@unist.ac.kr} \\
\normalsize \url{https://eu-ea.github.io}
}

\begin{document}
\maketitle
\begin{figure*}[t]
    \centering
    \includegraphics[width=0.9\textwidth]{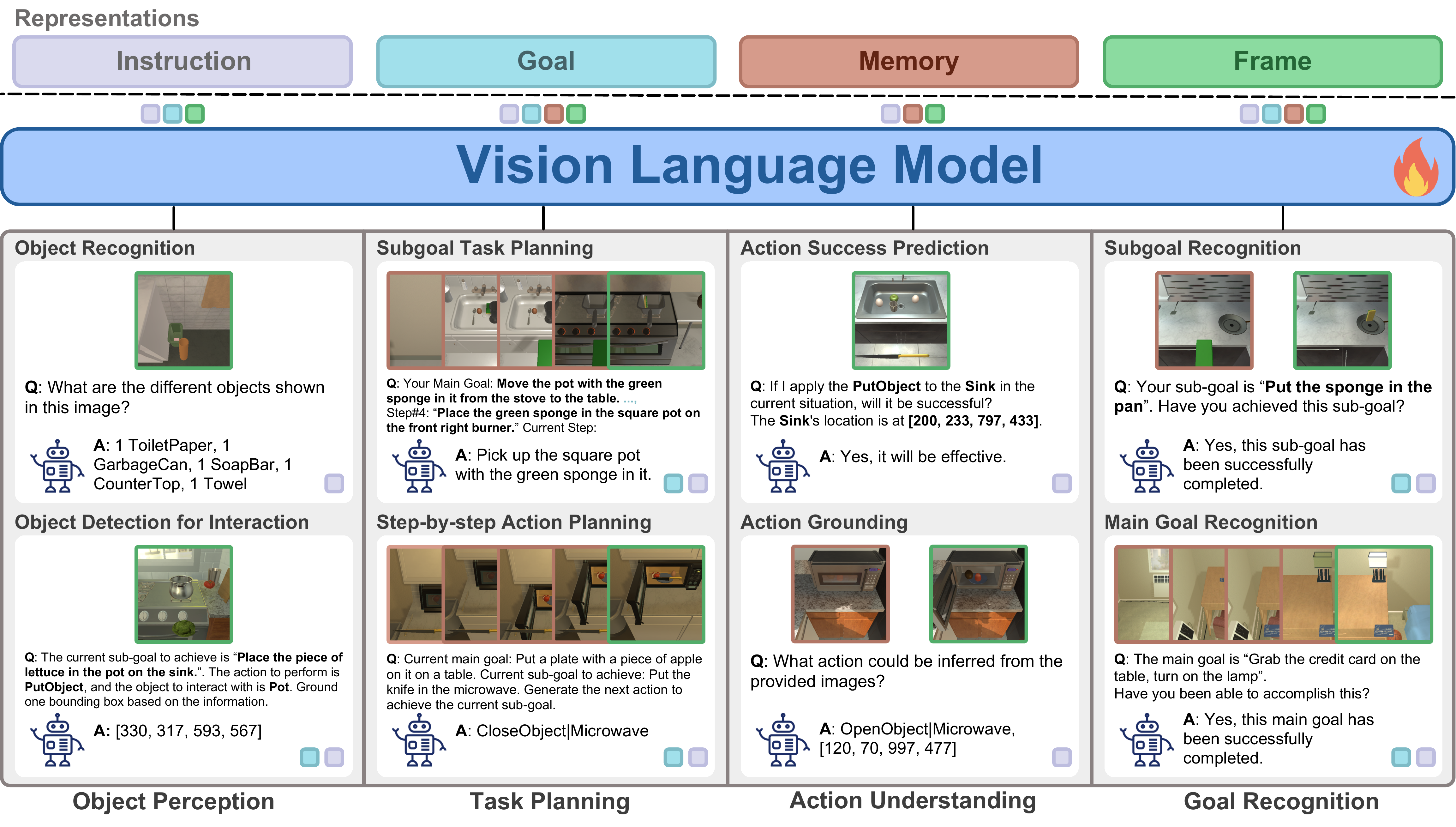}
    \vspace{-0.3cm}
    \caption{\textbf{Overview of {the four core skills of EUEA} to enhance VLM's environmental understanding and interaction.} Each core skill consists of two sub-skills, and {we fine-tune the VLM in a single stage using the data from all skills.} The colored boxes in each skill example indicate the representations. Additional each skill's examples template can be found in the supplementary material, Sec. B.}
    \label{fig:1}
    \vspace{-0.5cm}
\end{figure*}
\begin{abstract}
Vision-language models (VLMs) have shown strong perception and reasoning abilities for instruction-following embodied agents. However, despite these abilities and their generalization performance, they still face limitations in environmental understanding, often failing {on} interactions or relying on environment metadata during execution. To address this challenge, we propose a novel framework named Environmental Understanding Embodied Agent (EUEA), which fine-tunes four core skills: 1) \emph{object perception} for identifying relevant objects, 2) \emph{task planning} for generating interaction subgoals, 3) \emph{action understanding} for judging success likelihood, and 4) \emph{goal recognition} for determining goal completion. By fine-tuning VLMs with EUEA skills, our framework enables more reliable task execution for instruction-following. We further introduce a recovery step {that leverages these core skills} and a group relative policy optimization (GRPO) stage that {refines inconsistent skill predictions}. The recovery step samples alternative actions to correct failure cases, and the GRPO stage refines inconsistent skill predictions. Across ALFRED tasks, our VLM significantly outperforms a behavior-cloning baseline, achieving an 8.86\% improvement in average success rate. The recovery and GRPO stages provide an additional 3.03\% gain, further enhancing overall performance. Finally, our skill-level analyses reveal key limitations in the environmental understanding of closed- and open-source VLMs and identify the capabilities necessary for effective agent–environment interaction.
\end{abstract}
\vspace{-0.6cm} 
\section{Introduction}
\label{sec:introduction}

Environmental understanding is a core capability for embodied agents, enabling them to perceive, interpret, and interact with the environment to achieve {given tasks}. Recently, large language models (LLMs) and vision-language models (VLMs) have demonstrated impressive performance in the field of embodied agents \cite{ahn2022can, liang2022code, wang2023voyager, szot2025multimodal, chen2025training}, showing strong understanding and reasoning capabilities. However, despite their demonstrated abilities and generalization performance, they still face limitations in achieving environmental understanding. {Some prior embodied agent approaches~\cite{inoue2022prompter, zhao2024epo} rely on complex modular pipelines that require separate components, and other work~\cite{yang2024embodied} depends on environment metadata (e.g., object IDs, masks) to execute actions. On the other hand, end-to-end models often lack explicit environmental interpretation capabilities~\cite{szot2023large, szot2024grounding, szot2025multimodal}. As a result, execution errors lead to task failures. To overcome these failures, existing recovery methods rely on the environment to automatically validate outcomes \cite{sun2023adaplanner} and predefined failure types \cite{duan2024aha}, or they depend on textual feedback \cite{shinn2023reflexion}, which lacks the visual grounding essential for embodied agents.}

{To address these limitations, we propose the Environmental Understanding Embodied Agent (EUEA), a novel framework that bridges explicit skill modeling with end-to-end learning. Unlike approaches requiring complex module combinations, EUEA internalizes four core skills within a single VLM, without separate module modeling:} 1) \emph{object perception} for identifying relevant objects, 2) \emph{task planning} for generating interaction subgoals, 3) \emph{action understanding} for judging success likelihood, and 4) \emph{goal recognition} for determining goal completion. {This explicit skill-level supervision not only enables the model to perform visual perception and decision-making within a single architecture but also provides interpretability of its capabilities. Leveraging this unified formulation, we introduce a sampling-based recovery step without additional training that corrects failed interactions. Furthermore, we propose a group relative policy optimization (GRPO) \cite{shao2024deepseekmath} stage that refines inconsistent skill predictions via internal skill reward functions, thereby improving interaction performance.}

Embodied agents require navigation and interaction; the latter is particularly challenging due to reliance on hand-engineered components \cite{levine2016end}, making it difficult to generalize. By integrating strong VLM perception and reasoning with our core skills, EUEA enables end-to-end interaction for instruction-following. Our skill-based experiments also reveal limitations in existing closed- and open-source VLMs \cite{li2024llava, internvl2_5, internvl3, bai2025qwen2} and highlight the capabilities necessary for effective agent-environment interaction in instruction-following. In summary, our contributions are as follows:
\begin{itemize}
    \item We propose a {novel} EUEA framework that enhances the VLM's environmental understanding by integrating four core skills {without separate module modeling. Through this unified, end-to-end learning approach}, we enable the VLM to perceive, make decisions, and perform tasks, {thereby effectively improving interaction performance.}
    
    \item {We further introduce a sampling-based recovery step that leverages EUEA skills and a GRPO refinement stage.} The recovery step uses sampling to recover failure actions, and the GRPO stage refines inconsistent skill predictions. These provide an additional 3.03\% improvement in average task success on ALFRED \cite{shridhar2020alfred}, beyond the 8.86\% gain from supervised fine-tuning.
    
    \item We provide the skill datasets of 1.24M and 3.7M total samples and the skill evaluation benchmark using ALFRED \cite{shridhar2020alfred} and LangR \cite{szot2023large}, along with a method for constructing skill datasets. Additionally, we evaluate the skill performance of existing VLMs and demonstrate that our approach achieves superior environmental understanding.
\end{itemize}

\vspace{-0.1cm}
\section{Related Work}
\label{sec:related_work}
\vspace{-0.1cm}
\noindent\textbf{Environmental Understanding.} Environmental understanding is fundamental for successful embodied tasks \cite{zhou2024embodied, li2025embodied, yoo2025exploratory, intelligence2025pi05visionlanguageactionmodelopenworld}. Prior works has enhanced this capability through object-based representations \cite{hu2019you} and auxiliary modules \cite{zhu2020vision}. In vision-language navigation, detection-based scene representations \cite{chaplot2020object}, episodic memory for exploring area \cite{blukis2022persistent}, semantic voxel maps, and multi-view 3D reconstruction \cite{liu2024volumetric} improve spatial reasoning and navigation accuracy. More recent works \cite{pantazopoulos2023multitask} also identifies objects using visual tokens and defines action generation as skills, though they depend on separate object predictors and generate only the next action, without the ability to anticipate future state changes. Despite these advances, prior approaches \cite{huang2022language, song2023llm, choi2024lota, yang2025embodiedbench} generally improve environmental understanding through few-shot examples. {$\pi_{0.5}$} \cite{intelligence2025pi05visionlanguageactionmodelopenworld} performs instruction-following in real environments through open-world generalization, but it lacks {explicit environmental interpretation capabilities}. In contrast, our EUEA framework defines the entire instruction-following process as a set of skills based on a partially observable Markov decision process (POMDP)~\cite{kaelbling1998planning}, and integrates them into a single VLM via fine-tuning. This enables the VLM to perform end-to-end learning and to handle various functions such as perception, goal recognition, action planning, and future state anticipation in a unified structure, without requiring complex module combinations. Through this unified design, our method provides an integrated pathway toward robust environmental understanding.

\noindent\textbf{Interaction-Level Correction and Refinement.} Recent works has explored improving embodied agents by correcting errors during interaction through action-level repair \cite{duan2024aha} or planning-time refinement \cite{yang2025embodiedbench}. Existing methods recovers from failure by modifying the generated code using an inputted reason inferred from a fine-tuned model \cite{duan2024aha}, detect errors using self-feedback without visual grounding \cite{shinn2023reflexion}, or recover the interaction at the subgoal level using fixed plans \cite{zhang2021hierarchical}, symbolic state verification \cite{zhang2022danli}. Other LLM-based planning methods refine the error by environment feedback \cite{sun2023adaplanner}. While these methods refine the errors, they generally depend on predefined failure types \cite{duan2024aha}, external environment feedback \cite{sun2023adaplanner}, detects failure using self-feedback based on action executability but it does not consider visual information, thus cannot determine whether the action was actually successful \cite{wang2024e2cl}. In contrast, our approach recovers failed actions through sampling and further refines the model’s responses in the GRPO \cite{shao2024deepseekmath} stage, generating alternative or more confident action selections. This {self-correction mechanism} enables recovery without predefined failure types or external feedback, relying instead on the model’s learned environmental understanding. While prior studies \cite{shao2024deepseekmath, yu2025dapo} have primarily applied group-based reinforcement learning (RL) to enhance reasoning ability, we employ it to further correct actions in our framework. Recent work \cite{feng2025group} has also utilized group-based RL to stabilize multi-turn LLM agents. Inspired by this work, we adopt a GRPO refinement stage to stabilize EUEA skills and further improve task performance. Our method not only achieves performance gains without additional training via recovery step, but also yields further improvements through additional reinforcement. 
\vspace{-0.15cm}
\section{Method}
\vspace{-0.15cm}
\label{sec:method}
Inspired by embodied datasets and benchmarks \cite{embodiedscan, islam2024eqamx, Majumdar_2024_CVPR, mfe-etp, li2025embodied}, we focus on {equipping VLMs with grounded environmental understanding for instruction-following tasks}. {To achieve this, we define four core skills that are necessary for step-by-step interaction {based on a reward-free POMDP~\cite{kaelbling1998planning}}. At each time step $t$, the VLM receives only a single image frame $f_t$ and its accumulated past memory $\mathcal{M}$. Inferring all task-relevant information, such as visible objects $v_t$ from this partial view, the VLM selects the current action $a_t \in \mathcal{A}$, an interaction target $o_t$, and a bounding box $b_t$ to achieve the main goal $g \in \mathcal{G}$. After the action is executed, the entire interaction is recorded as a new memory state $m_t=(f_t,v_t,p_t,a_t,o_t,b_t,r_t,sg_t)$}. The memory $\mathcal{M}$ consists of these observations over time and includes a total of $T$ steps, represented as $\mathcal{M}=\{m_1, m_2, \dots, m_T\}, m_t\in \mathcal{M}, \forall t \in\{1, 2, \dots, T\}$. {Here, $p_t$ is the agent’s position, $r_t$ indicates the result of action at step $t$ including succeeded or failed, and $sg_t$ is the subgoal corresponding to step $t$. Examples from our EUEA skills are shown in Figure \ref{fig:1}.} We additionally introduce a recovery step via sampling, {leveraging EUEA skills to enable} the VLM to recover from failures. In Sec. \ref{sec:3-3}, we further introduce a GRPO {stage that refines inconsistent skill predictions.}

\subsection{Environmental Understanding Skills}
\vspace{-0.1cm}
The instruction, denoted as \(I_{\text{SKILL}}\), is tailored for each skill to produce the intended outcome and includes eight distinct skills: object recognition ($I_{OR}$), object detection ($I_{OD}$), subgoal task planning ($I_{STP}$), step-by-step action planning ($I_{SAP}$), action success prediction ($I_{ASP}$), action grounding ($I_{AG}$), main goal recognition ($I_{GR_{main}}$), and subgoal recognition ($I_{GR_{sub}}$). These instructions are provided in the supplementary material, Sec. B. 

\noindent\textbf{Object Perception.} {The object} perception skill consists of object recognition (OR) and object detection (OD), both of which support interaction. OR identifies objects present in the observed image, while OD detects a bounding box for an object relevant to a given current goal. These skills allow {a} VLM to recognize and localize them with a bounding box, enabling low-level control interaction. OR and OD are respectively formulated by the following equations:
\vspace{-0.15cm}
\begin{subequations}
\begin{align}
    v_t &= \pi_{\theta}(I_{OR}, f_t) \\
    b_t &= \pi_{\theta}(I_{OD}, sg_t, a_t, o_t, f_t)
\end{align}
\end{subequations}
Where $\pi_\theta$ denotes the VLM-based agent, and $b_t$ refers to the bounding box for the object $o_t$.

\noindent\textbf{Task Planning.}  Task planning is divided into subgoal task planning (STP) and step-by-step action planning (SAP) skills. STP generates subgoals to achieve a given main goal, while SAP generates an action to achieve a given current goal. These skills enable the VLM not only to identify the given main goal and plan interaction subgoals, but also to generate actions by taking $k$ past memory information, allowing it to interact with the environment as a performer to achieve the task. These skills formulated by the following equations:
\vspace{-0.15cm}
\begin{subequations} \label{eq:3-4}
\begin{align}
    sg_n &= \pi_{\theta}(I_{STP}, \mathcal{M}) \label{eq:3} \\
    a_t, o_t &= \pi_{\theta}(I_{SAP}, f_t, v_t, p_t, sg_t, m_{t-k:t-1}) \label{eq:4}
\end{align}
\end{subequations}

Where $sg_n$ denotes the $n$th subgoal generated by STP. 

\begin{figure}[t]
  \centering
    \includegraphics[width=0.99\linewidth]{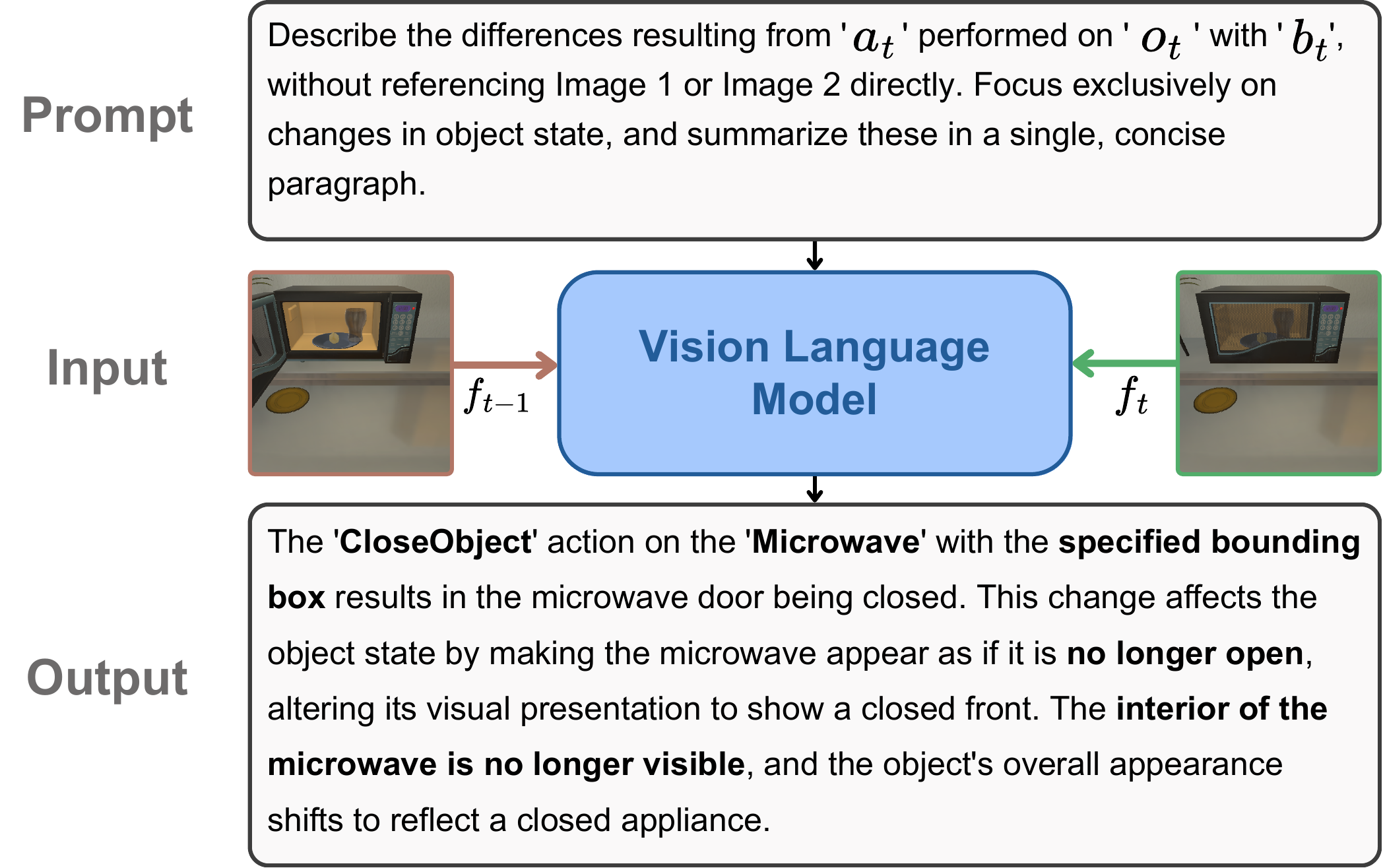}
    \caption{\textbf{Approach for generating data for environmental understanding.} We construct future situation captioning dataset that enables the prediction of captions describing changes between two images $f_{t-1}$ and $f_t$, from a single image $f_{t-1}$.}
   \label{fig:2}
\end{figure}

\noindent\textbf{Action Understanding.} We assume that meaningful environmental understanding requires a VLM to anticipate the outcomes of its actions and to describe the resulting state changes in the environment. To support this capability, we design three skills: action success prediction (ASP), future situation captioning (FSC), and action grounding (AG). ASP predicts whether an action will succeed or fail in the current situation. FSC is an extended ASP skill that allows VLM to describe the expected changes resulting from a given action. AG allows VLM to predict which action was executed between given two images. To build these skills, especially ASP and FSC, the dataset must contain sufficient failure examples, but expert demonstrations primarily include successful interactions. To address this imbalance, we generate failure-rich data through random exploration, where the VLM performs zero-shot random actions within the discrete action space. This process yields diverse successful and failed interactions across scenes. FSC data is additionally constructed by generating captions that explain the differences between pre- and post-action images using both expert trajectories and random-exploration memory as shown in the Figure \ref{fig:2}. These skills are designed to predict action success, allowing the VLM to implicitly learn the constraints of interacting with the environment, thereby enhancing its understanding of the environment. ASP, FSC, and AG are respectively formulated by the following equations:
\vspace{-0.1cm}
\begin{subequations} \label{eq:5-7}
\begin{align}
    ASP_t &= \pi_{\theta}(I_{ASP}, a_t, o_t, b_t, f_t) \label{eq:5} \\
    FSC_t &= \pi_{\theta}(I_{FSC}, a_t, o_t, b_t, f_t) \label{eq:6} \\
    a_{t-1}, o_{t-1}, b_{t-1} &= \pi_{\theta}(I_{AG}, f_{t-1:t}) \label{eq:7}
\end{align}
\end{subequations}

Where $ASP_t$ represents the predicted result of an action at step $t$, including "Yes" or "No", while $FSC_t$ denotes a caption describing the predicted outcome of the action at step $t$. $I_{FSC}$ is the instruction input to FSC.

\noindent\textbf{Goal Recognition.} Goal recognition is the skill that enables an agent to determine whether a subgoal or main goal has been achieved. The VLM learns to determine the achievement of the subgoal based on the main goal, or the achievement of the main goal, enhancing its understanding of the given task. $GR_{main}$ and $GR_{sub}$ are respectively formulated by the following equations:
\vspace{-0.1cm}
\begin{subequations} \label{eq:8-9}
\begin{align}
    GR_{main} &= \pi_{\theta}(I_{GR_{main}}, \mathcal{M}) \label{eq:8} \\
    GR_{sub} &= \pi_{\theta}(I_{GR_{sub}}, sg_t, a_{t-n:t}, f_{t-n:t}) \label{eq:9}
\end{align}
\end{subequations}

Where $GR_{main}$ represents the predicted result for the main goal, including "Yes" or "No", while $GR_{sub}$ similarly represents the predicted result for the subgoal, also including "Yes" or "No". $a_{t-n:t}$ and $f_{t-n:t}$ denote the actions and frames corresponding to the past $n$ steps of information and current step $t$ from the given subgoal $sg_t$, including step $t$. This formulation allows the VLM to recognize both its ability to accomplish the overarching main goal and its progress  in achieving intermediate subgoals, thereby improving overall task understanding. In particular, this $GR_{sub}$ enables the model to decide when to move to the next subgoal.

\noindent\textbf{Skill Dataset Construction.} We use two instruction-following benchmarks to construct our skill datasets: ALFRED \cite{shridhar2020alfred}, built on AI2-THOR \cite{kolve2017ai2}, and LangR rearrangement benchmark \cite{szot2023large}, built on Habitat 2.0 \cite{szot2021habitat}. ALFRED consists of 8,055 expert demonstrations, 25,743 human annotations, and 428k image-action pairs, while LangR includes 150k episodes and 55k instructions. From these benchmarks, we collect environmental observations from image-action pairs and annotations, and design prompt templates to construct the four core skill datasets that support environmental understanding. To assess generalization to unseen environments, we additionally sample 31 of the 108 ALFRED training scenes and 42 of the 84 LangR training scenes. Using this data, we build four skill datasets containing 382k training, 858k validation, and 5.3k evaluation samples for ALFRED, and 906k training, 2.8M validation, and 4.4k evaluation samples for LangR. Across ALFRED and LangR, the fine-tuning sets include 108k and 237k samples for object perception, 25k and 144k for task planning, 217k and 471k for action prediction, and 29k and 52k for goal recognition. The evaluation sets contain 1k and 1k samples for object perception, 1.2k and 2k for task planning, 2.1k and 1k for action prediction, and 1k and 0.5k for goal recognition. The full data templates and collection pipeline are detailed in the supplementary material, Sec. B.

\subsection{Recovery Step via Sampling}
Despite utilizing fine-tuned skills such as step-by-step action planning (SAP) and object detection (OD) to interact with the environment, interaction failures can still occur. To address this, we introduce a \textbf{Recovery Step}, which is triggered when action fails. In this step, the VLM agent $\pi_{\theta}$ samples alternative actions using SAP skill. The sampled output differs from the failed one but remains aligned with the given goal, guided by the corresponding instruction $I_{SAP}$. This alternative actions are selected based on the score equation:
\vspace{-0.1cm}
\begin{equation} \label{eq:10}
s_{t,i} = -\log \pi_{\theta}\!\left(a_{t,i},o_{t,i} \mid I_{SAP}, m_{t-k:t}\right)
\end{equation}
Where $i$ is sampling index, and the score $s_{t,i}$ is the negative log-likelihood of the SAP skill from Equation \ref{eq:4}. Since the $\pi_{\theta}$ generates actions conditioned on the given goal, the predicted probability can be interpreted as goal-achievement confidence, allowing $\pi_{\theta}$ to select the most promising action for the given goal. 

To recover from failure, we conduct $n$ SAP to generate new candidate actions $a_{t,i}$ with objects $o_{t,i}$. When $\pi_{\theta}$ assigns a high probability to a sampled action-object pair, the score $s_{t,i}$ is close to zero. If all sampled action–object pairs correspond to the same previously failed action, we instead perform OD $n$ times and compute the score $s_t$ for the previously failed action–object pair using Equation \ref{eq:10} similarly. After sampling $n$ times, $\pi_{\theta}$ selects either an action-object pair $a_{new}$, $o_{new}$ or a bounding box $b_{new}$ with the lowest score to interact with the environment. If an action-object pair is selected, we perform OD to obtain the corresponding bounding box $b_{new}$. This step ensures recovery from two types of failure scenarios  through sampling without additional training.

\vspace{-0.1cm}
\subsection{GRPO Refinement Stage} \label{sec:3-3}
\vspace{-0.1cm}
Although supervised fine-tuning (SFT) on the our four skills improves environmental understanding, the VLM can still produce incorrect decisions. To address this issue, we introduce a GRPO \cite{shao2024deepseekmath} refinement stage that reduces inconsistent responses through rule-based rewards. This framework enables the VLM to further enhance its understanding of the environment.

\noindent\textbf{Reward Function.} We define reward functions using task-specific correctness metrics, including metrics based on intersection of union (IoU), the Jaccard index, and action-sequence order. The total reward function is given by: $R_{total} = R_{OP}+R_{TP}+R_{AU}+R_{GR}$, where $R_{OP}$, $R_{TP}$, $R_{AU}$, and $R_{GR}$ are the reward functions for the four core skills: object perception ($R_{OP}$), task planning ($R_{TP}$), action understanding ($R_{AU}$), and goal recognition ($R_{GR}$). {Since each input instance corresponds to a single skill, the rewards for all other skills are set to zero.} $R_{OP}$ uses the Jaccard index to reward correct predictions for the OR subskill and applies an IoU based reward for bounding boxes in OD. $R_{TP}$ assigns rewards based on the correctness of the predicted action-object pair in SAP and the correctness of the predicted action sequence order in STP. $R_{AU}$ rewards correct success predictions in ASP and FSC. In AG, it combines the correctness of the action-object pair used in SAP and OD with an IoU weighted reward for the bounding box. Finally, $R_{GR}$ provides rewards based on the correctness of both main goal and subgoal predictions. The detailed reward values and the complete formulation of all reward functions are provided in supplementary material, Sec. C.

\noindent\textbf{Dataset construction and Strategy.} We use the previously created validation set to collect eight response samples for all data and then filter out the ambiguous cases. Our sampling strategy proceeds as follows: 1) Sampling eight responses for each data instance. 2) Counting the number of correct responses $c$ for each instance. 3) Selecting instance whose normalized standard deviation of rewards, exceeds a threshold $\tau$. This allows us to construct a compact dataset of around 10k instances on ALFRED \cite{shridhar2020alfred}, consisting only of cases where the model shows uncertainty, without using the entire validation set. Even with this smaller but compact dataset, this stage improve the VLM’s decision-making ability by refining ambiguous responses. 
\vspace{-0.2cm}
\section{Experiments}
\label{sec:experiments}
\vspace{-0.1cm}

\subsection{Experimental Setup} \label{sec:4_1}
\textbf{Training Details.} We use InternVL3-8B \cite{internvl3} as our main VLM for ALFRED \cite{shridhar2020alfred} and LangR \cite{szot2023large}, due to their strong multi-modal understanding, scalability, and support for multiple image inputs. To assess the contribution of our environmental understanding skills, we also train a behavior cloning (BC) baseline in which all skills are removed except OD, SAP, $GR_{sub}$. In SFT stage, we full fine-tune all VLMs for 1 epoch each using 8 A100 80GB GPUs, following the their training scripts \cite{internvl2_5, internvl3, zheng2024llamafactory}. The vision encoder is frozen, while the MLP and LLM components are fine-tuned. We set the batch size to 128, the max sequence length to 8192. In the GRPO stage, we fine-tune InternVL3-8B with LoRA \cite{hu2022lora} for 5 out of 10 epochs due to early stopping, using 2 A100 80GB GPUs, following DAPO \cite{yu2025dapo} hyperparameter setting. We set the batch size to 64 and the max sequence length to 8192.

\noindent\textbf{Task Evaluation.} To evaluate whether the learned skills improve VLM’s task performance, we evaluate our models on ALFRED \cite{shridhar2020alfred}, which provides clearly defined state changes for robust task evaluation, including long-horizon interaction tasks. We follow ALFWorld \cite{shridhar2020alfworld} by using 134 instruction-following tasks grounded in ALFRED states, and, motivated by EMMA \cite{yang2024embodied} which emphasizes the value of human-annotated free-form instructions, we augment them with additional free-form variants, yielding 429 evaluation tasks in total. These tasks are categorized into six type: Examine in Light (Look), Pick\&Place (Pick), Pick Two\&Place (Pick Two), Clean\&Place (Clean), Cool\&Place (Cool), and Heat\&Place (Heat). We exclude LangR \cite{szot2023large} from task evaluation because its trajectories often include interactions with non-visible objects, creating scenarios that are incompatible with our setting where the agent may act only on visible observations. We construct an instruction-following pipeline where VLM takes only instructions and observed images from the environment as inputs, while storing and utilizing intermediate outputs in memory to perform sequentially. We categorize the given subgoals into two types: \textbf{Navigation} and \textbf{Interaction}. In the \textbf{Navigation} phase, the agent follows PDDL expert actions while gathering environmental information through $OR$. During the \textbf{Interaction} phase, the agent generates actions to achieve subgoals through $SAP$, subsequently predicts the target object and its corresponding bounding box through $OD$, and executes the interaction with the environment based on these predictions. We treat an action as failed when the image does not change, which fits discrete-action settings with distinct visual transitions. This cycle continues until all given subgoals are completed, using $GR_{sub}$. We set $k$ to 4 for the memory input and $n$ to 10 for the recovery step.

\begin{figure*}[t]
    \centering
    \begin{minipage}{0.6\textwidth}
        \captionsetup{type=table} 
        \caption{\textbf{Comparison of task success rates with instruction-following VLM Agents.} * indicates that they are from the reported results \cite{shridhar2020alfred, yang2024embodied}.}
        \label{tab:1}
        \vspace{-0.2cm}
        \resizebox{0.99\linewidth}{!}{
        \begin{tabular}{
            c@{\hspace{2.5pt}}l c@{\hspace{2.5pt}}c lr@{\hspace{2.5pt}}lr@{\hspace{2.5pt}}lr@{\hspace{2.5pt}}lr@{\hspace{2.5pt}}lr@{\hspace{2.5pt}}lr@{\hspace{2.5pt}}l}
            \toprule
            \multicolumn{2}{l}{\multirow{2}{*}{\textbf{VLM Agent}}} & \textbf{} & \multicolumn{14}{c}{\textbf{Task Success Rate}} \\
            \cline{4-17}
            & & \textbf{} & \multicolumn{2}{c}{Avg.} & \multicolumn{2}{c}{Look} & \multicolumn{2}{c}{Pick} & \multicolumn{2}{c}{Pick Two} & \multicolumn{2}{c}{Clean} & \multicolumn{2}{c}{Cool} & \multicolumn{2}{c}{Heat} \\
            \midrule 
            
            \multicolumn{2}{l}{EMMA* \cite{yang2024embodied}}             & &\multicolumn{2}{c}{67.83} &\multicolumn{2}{c}{66.67} &\multicolumn{2}{c}{71.95} &\multicolumn{2}{c}{75.93} &\multicolumn{2}{c}{{65.31}} &\multicolumn{2}{c}{55.56} &\multicolumn{2}{c}{71.80} \\
    
            \multicolumn{2}{l}{{Human Performance*~\cite{shridhar2020alfred}}} & &\multicolumn{2}{c}{{91.00}} &\multicolumn{2}{c}{-} &\multicolumn{2}{c}{-} &\multicolumn{2}{c}{-} &\multicolumn{2}{c}{-} &\multicolumn{2}{c}{-} &\multicolumn{2}{c}{-} \\
            \midrule
    
            \multicolumn{2}{l}{BC (InternVL3-8B)} & &\multicolumn{2}{c}{74.59} &\multicolumn{2}{c}{\underline{88.89}} &\multicolumn{2}{c}{{73.17}} &\multicolumn{2}{c}{{57.41}} &\multicolumn{2}{c}{62.24} &\multicolumn{2}{c}{\underline{96.83}} &\multicolumn{2}{c}{{75.64}} \\
    
            \multicolumn{2}{l}{Ours (SFT)} & &\multicolumn{2}{c}{\underline{83.45}} &\multicolumn{2}{c}{\textbf{90.74}} &\multicolumn{2}{c}{\textbf{86.59}} &\multicolumn{2}{c}{\underline{75.93}} &\multicolumn{2}{c}{\underline{65.31}} &\multicolumn{2}{c}{\textbf{98.41}} &\multicolumn{2}{c}{\textbf{91.03}} \\

            \multicolumn{2}{l}{Ours (GRPO)} & &\multicolumn{2}{c}{\textbf{85.78}} &\multicolumn{2}{c}{\textbf{90.74}} &\multicolumn{2}{c}{\underline{85.37}} &\multicolumn{2}{c}{\textbf{85.19}} &\multicolumn{2}{c}{\textbf{74.49}} &\multicolumn{2}{c}{\textbf{98.41}} &\multicolumn{2}{c}{\underline{87.18}} \\
    
            \bottomrule
        \end{tabular}}
    \end{minipage}
    \hspace{5pt}
    \centering
    \begin{minipage}{0.375\textwidth}
        \captionsetup{type=table}
        \caption{{\textbf{Comparison of recovery methods for task evaluation.} We evaluate task performance by comparing different recovery methods. \emph{Env Feedback} adds external environment feedback when it failed.}}
        \vspace{-0.2cm}
        \resizebox{0.99\linewidth}{!}{
        \begin{tabular}{llcccc}
            \toprule
            \multicolumn{2}{l}{\multirow{2}{*}{\textbf{Recovery Method}}} & \multicolumn{2}{c}{\textbf{Success Rate}} & \multicolumn{2}{c}{\textbf{Goal Condition}} \\
            \cmidrule(lr){3-4} \cmidrule(lr){5-6}
            & & SFT & GRPO & SFT & GRPO \\
            \midrule
            \multicolumn{2}{l}{Ours} & 83.45 & 85.78 & 88.42 & 90.17 \\
            \multicolumn{2}{l}{w/ Env feedback} & \textbf{85.78} & 85.78 & \textbf{90.09} & 90.17 \\
            \multicolumn{2}{l}{w/ Recovery Step} & \textbf{85.78} & \textbf{86.48} & 89.74 & \textbf{90.48} \\
            \bottomrule
        \end{tabular}}
       
        \label{tab:2}
    \end{minipage}
    \vspace{-0.2cm}
\end{figure*}

\begin{table*}[t]
\centering
\caption{\textbf{Skill evaluation results of closed- and open-source VLMs.} We evaluate both closed- and open-source models using the evaluation dataset generated from four skills utilizing ALFRED \cite{shridhar2020alfred} and LangR \cite{szot2023large}. \textbf{Detection} is measured using IoU, \textbf{Planning} is evaluated using a BERT-based transformer \cite{sbert} with cosine similarity between subgoals, and all other skills are evaluated based on accuracy. Navigation* represents three additional sub-skills for navigation. Ours* indicates the InternVL3-8B VLM fine-tuned with ALFRED skills.}
\label{tab:3}
\vspace{-0.225cm}
\begin{adjustbox}{width=\textwidth}
\begin{tabular}{clcccccccc}
    \toprule
    \multicolumn{1}{c}{\multirow{2}{*}{\textbf{ALFRED}}}&\multicolumn{1}{l}{\multirow{2}{*}{\textbf{Model}}} & \multicolumn{2}{c}{\textbf{Object Perception}} & \multicolumn{2}{c}{\textbf{Goal Recognition}} & \multicolumn{2}{c}{\textbf{Action Understanding}} & \multicolumn{2}{c}{\textbf{Task Planning}} \\
    \cmidrule(lr){3-4} \cmidrule(lr){5-6} \cmidrule(lr){7-8} \cmidrule(lr){9-10}
    & & Grounding & Detection & Main & Sub & Prediction & Grounding & Planning & Step-by-step \\
    \midrule
    
    \multirow{5}{*}{\rotatebox[origin=c]{0}{\shortstack{Closed}}}
    & GPT-5       & 51.45 & 24.34 & 92.20 & 73.80 & 81.48 & {28.86} & {0.801} & 71.52 \\
    & GPT-o3      & {57.28} & 29.55 & \underline{94.80} & {80.60} & \underline{86.75} & 31.76 & 0.796 & {77.41} \\
    & Claude-4.5-Sonnet   & {38.05} & 1.54 & {90.80} & {65.80} & {51.60} & 62.55 & 0.812 & {46.15} \\
    & Gemini-2.5-Flash    & {45.00} & 43.15 & {88.60} & {66.80} & {69.45} & 32.53 & 0.799 & {74.80} \\
    & Gemini-2.5-Pro      & {63.53} & 60.75 & {86.20} & {80.20} & {85.43} & 31.08 & \underline{0.819} & {85.76} \\
    \midrule
    
    \multirow{5}{*}{\rotatebox[origin=c]{0}{\shortstack{Open}}}
    &LLaVA-OneVision-7B~\cite{li2024llava} & 22.19 & {18.19} & 69.60 & 68.00 & 71.05 & 44.88 & 0.695 & 6.87 \\
    &InternVL2.5-8B~\cite{internvl2_5} & 30.90 & 8.79 & 67.00 & 76.20 & 68.70 & 47.39 & 0.608 & 0.82 \\
    &InternVL3-8B~\cite{internvl3} & 33.70 & 26.68 & 71.80 & 77.40 & 68.80 & 48.26 & 0.742 & 4.42 \\
    &Qwen2.5-VL-7B~\cite{bai2025qwen2} & 28.15 & 1.82 & 84.00 & 73.00 & 48.78 & 49.61 & 0.764 & 39.44 \\
    &Qwen3-VL-8B~\cite{bai2025qwen3} & 48.99 & 51.79 & 89.40 & 80.20 & 76.50 & \underline{67.18} & 0.815 & 45.34 \\
    \midrule
    & BC (InternVL3-8B) & \textbf{78.01} & \underline{73.94} & {71.80} & \underline{83.00} & {63.25} & {9.27} & {0.630} & \textbf{98.53} \\
    \midrule
    & Ours (InternVL3-8B) & \underline{75.84} & \textbf{81.73} & \textbf{99.40} & \textbf{98.60} & \textbf{96.80} & \textbf{89.09} & {\textbf{0.894}} & {\underline{98.20}} \\
    \bottomrule
    \multicolumn{1}{c}{\multirow{2}{*}{\textbf{LangR}}}&\multicolumn{1}{l}{\multirow{2}{*}{\textbf{Model}}} & \multicolumn{2}{c}{\textbf{Object Perception}} & \multicolumn{2}{c}{\textbf{Goal Recognition}} & \multicolumn{2}{c}{\textbf{Action Understanding}} & \multicolumn{2}{c}{\textbf{Task Planning}} \\
    \cmidrule(lr){3-4} \cmidrule(lr){5-6} \cmidrule(lr){7-8} \cmidrule(lr){9-10}
    & & Grounding & Detection & \multicolumn{2}{c}{Main} & Prediction & Grounding & Navigation$^*$ & Step-by-step \\
    \midrule
    \multirow{5}{*}{\rotatebox[origin=c]{0}{\shortstack{Closed}}}
    & GPT-5 & 35.07 & {12.35} & \multicolumn{2}{c}{87.40} & 62.50 & 28.29 & 60.53 & 88.02 \\

    & GPT-o3 & 35.17 & {20.77} & \multicolumn{2}{c}{95.80} & 72.96 & 27.30 & 59.67 & 85.57 \\
    & Claude-4.5-Sonnet & 24.96 & 0.00 & \multicolumn{2}{c}{64.40} & 38.27 & 21.50 & 43.80 & 89.00 \\
    & Gemini-2.5-Flash & 30.43 & 24.08 & \multicolumn{2}{c}{94.40} & 71.94 & 27.55 & 59.20 & 77.26 \\

    & Gemini-2.5-Pro & 45.68 & 59.77 & \multicolumn{2}{c}{\underline{95.40}} & \underline{82.14} & {30.35} & 38.80 & 97.07 \\
    \midrule
    \multirow{5}{*}{\rotatebox[origin=c]{0}{\shortstack{Open}}}
    &LLaVA-OneVision-7B~\cite{li2024llava} & 11.15 & {8.52} & \multicolumn{2}{c}{53.00} & 42.60 & 0.19 & 39.53 & 40.34 \\
    &InternVL2.5-8B~\cite{internvl2_5} & {25.99} & 1.12 & \multicolumn{2}{c}{75.00} & 44.00 & 12.59 & 47.53 & 30.32 \\
    &InternVL3-8B~\cite{internvl3} & {29.03} & 8.62 & \multicolumn{2}{c}{74.60} & 48.00 & 13.10 & 52.90 & 69.68 \\
    &Qwen2.5-VL-7B~\cite{bai2025qwen2} & 21.64 & 0.00 & \multicolumn{2}{c}{64.00} & 36.22 & 22.04 & 57.07 & {79.46} \\
    &Qwen3-VL-8B~\cite{bai2025qwen3} & 38.41 & {29.94} & \multicolumn{2}{c}{91.00} & 63.27 & 27.22 & 60.60 & 72.13 \\
    \midrule
    \multirow{1}{*}{\rotatebox[origin=c]{0}{\shortstack{Cross}}}
    & Ours$^{*}$ (ALFRED-tuned) & {26.81} & {58.35} & \multicolumn{2}{c}{{86.60}} & {77.81} & \underline{47.52} & {-} & {77.06} \\
    \midrule
    & BC (InternVL3-8B) & \textbf{86.59} & \underline{88.93} & \multicolumn{2}{c}{\textbf{100.00}} & {38.78} & {14.14} & \underline{73.67} & \underline{98.04} \\
    \midrule
    & Ours (InternVL3-8B) & \underline{86.50} & \textbf{93.20} & \multicolumn{2}{c}{\textbf{100.00}} & \textbf{100.00} & \textbf{49.84} & \textbf{74.07} & \textbf{99.27} \\
    \bottomrule
\end{tabular}
\end{adjustbox}
\vspace{-0.475cm}
\end{table*}

\begin{figure}[t]
  \centering
    \includegraphics[width=0.999\linewidth]{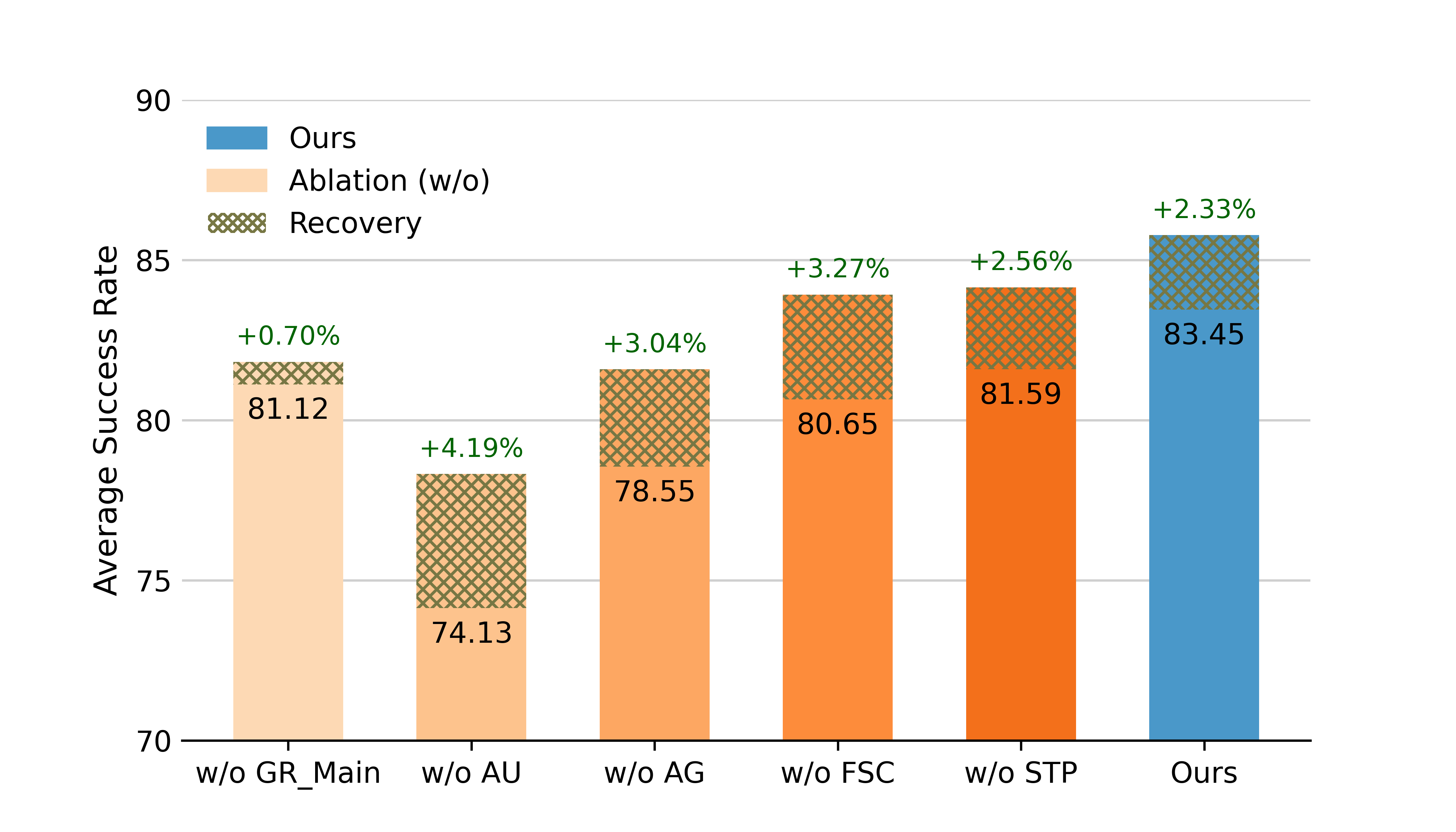}
    \vspace{-0.75cm}
    \caption{{\textbf{Ablation study on our proposed skills for task evaluation.} We evaluate task performance by conducting an ablation study on the proposed skills that impact environmental understanding. $AG$ refers to action grounding, $FSC$ to future situation captioning, $AU$ to action understanding, $STP$ to subgoal task planning, and $GR_{Main}$ to main goal recognition, respectively.}}
   \label{fig:3}
   \vspace{-0.5cm}
\end{figure}

\noindent\textbf{Skill Evaluation.} To evaluate how effective VLMs are in generating interactions for embodied AI, we propose an evaluation benchmark based on four key skills, each skill has two metrics: \textbf{1) Object Grounding} evaluates object identification accuracy based on count and inclusion correctness. \textbf{2) Object Detection} measures bounding box accuracy using Intersection over Union (IoU). \textbf{3) Planning} evaluates generated and ground-truth subgoals via cosine similarity with a BERT-based transformer \cite{sbert}. \textbf{4) Step-by-Step} checks if both the action and object match the ground truth. \textbf{5) Action Prediction} evaluates "Yes" or "No" response accuracy for action success. \textbf{6) Action Grounding} verifies whether the model correctly identifies action, object, and bounding box across images. \textbf{7) Goal Recognition (Main \& Sub)} evaluates yes/no response accuracy for goal completion. We evaluate our fine-tuning dataset generated with ALFRED \cite{shridhar2020alfred} and LangR \cite{szot2023large} using both open source VLMs \cite{li2024llava, internvl2_5, internvl3, bai2025qwen2} and closed state-of-the-art VLMs \cite{achiam2023gpt}. Additionally, in LangR, we define three sub-skills for navigation: 1) identifying where to go in order to perform actions such as "Pick" or "Place" in rearrangement tasks, 2) selecting the target object, and 3) selecting the destination. 

Due to space limits, we provide detailed descriptions of the sub-skills, evaluation prompts for all skills, and the pipeline algorithm in the supplementary material, Secs. B–E.

\subsection{Result} \label{sec:4_2}
\noindent\textbf{Task Evaluation.} Table \ref{tab:1} compares our VLMs with the BC baseline on task evaluation using ALFRED \cite{shridhar2020alfred}. Our VLMs show significant improvements on most tasks over the BC baseline, achieving average task-success gains of 8.86\% and 10.96\% in the SFT and GRPO stages, respectively. This indicates that the four core skills enhance environmental understanding, with the GRPO stage further refines inconsistent responses, where correct and incorrect predictions alternate, leading to even stronger environmental understanding. Although EMMA \cite{yang2024embodied} extends ALFWorld \cite{shridhar2020alfworld} to a vision-based setting, it relies on high-level interactions and environment metadata, and operates without subgoals. In contrast, our model depends solely on visual observations and low-level actions given subgoals, making the interaction setting more demanding while still achieving strong fine-grained performance.

\noindent\textbf{Comparison of Recovery Methods.} We conduct a comparative analysis to evaluate the effect of our recovery method on task performance compared to the baseline. The baseline recovery relies on additional external environment feedback after failures to guide recovery. We include this baseline for comparison, as prior works either utilize environmental feedback directly \cite{sun2023adaplanner}, train models through such feedback \cite{wang2024e2cl}, or define failure types \cite{duan2024aha} that can be attributed to environmental feedback. Table \ref{tab:2} shows that our recovery step achieves the highest task success rate for both SFT and GRPO stage, especially outperforming oracle feedback in GRPO stage. This indicates that, rather than relying on external environmental feedback, sampling based on the agent’s environmental understanding can achieve comparable or even superior performance.

\noindent\textbf{Skill Evaluation.} Table \ref{tab:3} presents the evaluation results of the skill datasets using ALFRED and the photorealistic LangR \cite{szot2023large} benchmarks. Across both benchmarks, Gemini-2.5-Pro achieves the strongest zero shot performance in most skills despite not being trained on either dataset. Its strong reasoning ability improves skill prediction, particularly for goal recognition and planning. In goal recognition, the performance gap between Gemini-2.5-Pro and GPT-o3, and our fine-tuned model is only 4.6\% in both skill benchmarks, which is consistent with prior works \cite{yao2023react, shinn2023reflexion, sun2023adaplanner, yang2025embodiedbench} showing the strong planning capabilities of LLMs and VLMs.

However, most zero-shot VLMs still lack environmental understanding, resulting in limited accuracy in action success prediction, action grounding, and object perception in both skill benchmarks. In ALFRED’s step-by-step action planning, Gemini-2.5-Pro reaches 85.76 percent accuracy, meaning that nearly 14\% of individual actions fail, which often leads to task failure. This challenge becomes even greater as the action space increases. In contrast, our model, fine tuned on the four proposed skills, significantly improves performance across all evaluations. By enhancing EUEA skills, it outperforms the baselines and demonstrates stronger performance on the skills that support instruction-following, even in real world settings. Although BC shows similar or slightly better results on some skills, these differences do not lead to stronger task performance. These results highlight the need for additional training to improve environmental understanding, ensuring VLMs function as agents.

\begin{figure}[t]
  \centering
    \includegraphics[width=0.99\linewidth]{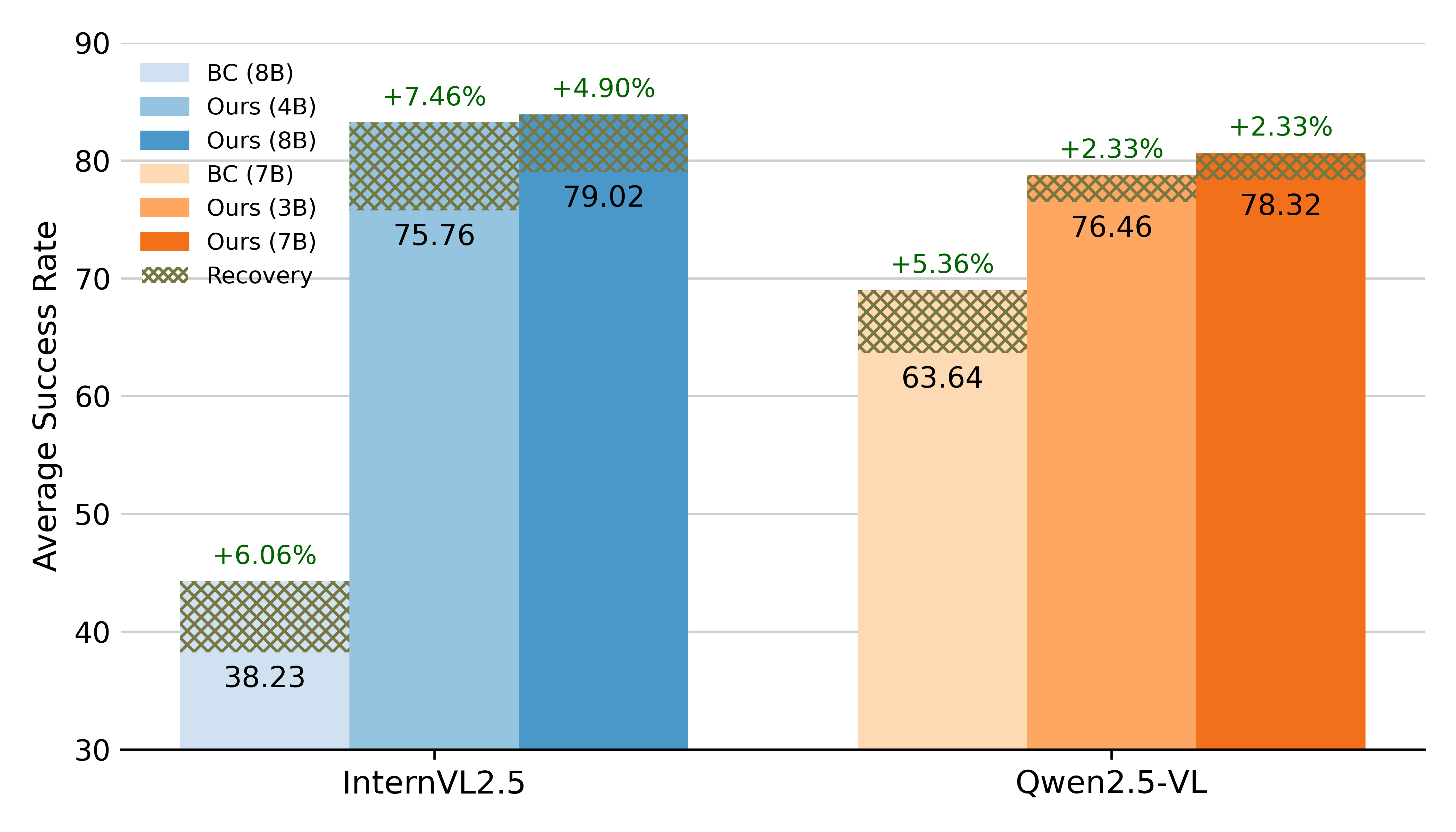}
    \vspace{-0.3cm}
    \caption{\textbf{Comparison of VLM backbones on task evaluation.} We evaluate task performance using the InternVL2.5-series and Qwen2.5-VL-series as backbones in SFT stage.}
   \label{fig:4}
   \vspace{-0.5cm}
\end{figure}

\begin{figure*}[t]
    \centering
    \includegraphics[width=0.905\textwidth]{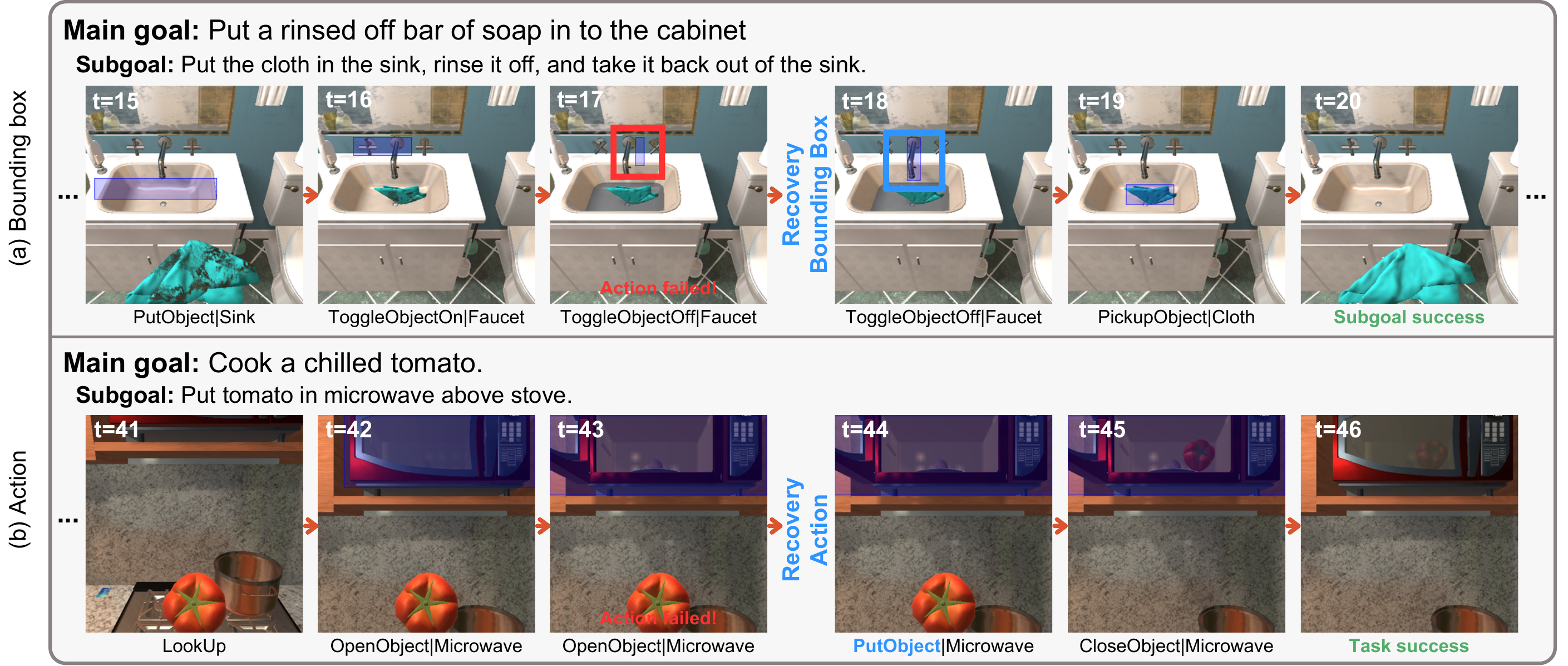}
    \vspace{-0.2cm}
    \caption{\textbf{Analysis of cases where the recovery step resolves a failed interaction.} (a) shows when an incorrect detection is corrected, allowing the action to succeed. (b) shows when an interaction fails, and an alternative action completes the given task successfully.} 
    \label{fig_qualitative}
    \vspace{-0.4cm}
\end{figure*}

\subsection{Analyses} \label{sec:4_3}
\noindent\textbf{Impact of Task Performance under Skill Ablation.} We analyze the impact of the proposed skills on task evaluation. Figure \ref{fig:3} presents an ablation study on main goal recognition ($GR_{Main}$), action understanding ($AU$), action grounding ($AG$), future situation captioning ($FSC$), and subgoal task planning ($STP$), which are not explicitly used in the evaluation pipeline. Removing each skill individually results in task success rate drops of 2.33\% for $GR_{Main}$, 9.32\% for $AU$, 4.9\% for $AG$, 1.86\% for $FSC$, and 2.8\% for $STP$, corresponding Ours SFT stage. These results show that removing any skill drops performance, with $AU$ having the largest impact. This demonstrates that each skill contributes meaningfully to task evaluation and that improved prediction of environment changes leads to higher task success rate.

\noindent\textbf{Analysis of Task Performance across VLM Backbones.} Figure \ref{fig:4} presents the task performance results of our method after SFT using different VLM backbones, InternVL2.5-series \cite{internvl2_5} and Qwen2.5-VL-series \cite{bai2025qwen2}. Our approach improves the average success rate over the respective BC baselines for both series. Furthermore, even with models that have about 50\% fewer parameters, InternVL2.5-4B and Qwen2.5-VL-3B, our method achieves notable performance gains of 12.84\% and 37.53\%, respectively, compared to their BC baselines. These results demonstrate that our approach effectively enhances average task performance and remains robust across different backbone scales.

\noindent\textbf{Out-of-Distribution (OOD) Skill Evaluation.} We evaluate the VLM fine-tuned on the ALFRED skill dataset using the LangR skill benchmark to evaluate cross-environment generalization. Although both datasets contain Pick\&Place interactions, ALFRED has more diverse action space, whereas LangR focuses only on simple, photorealistic Pick\&Place rearrangement scenes. To address vocabulary mismatches, we map actions and objects using a BERT-based transformer~\cite{sbert} (e.g., "Pick" to "PickupObject", "CoffeeTable" to "Table"). As shown in Table \ref{tab:3}, the ALFRED-tuned model performs substantially lower than the LangR-tuned model, which is expected given the distribution gap, but it still outperforms the zero-shot InternVL3-8B on {all skills except object grounding}. This result shows that the EUEA skills transfer even under significant distribution shift, indicating promising OOD generalization.

\noindent\textbf{Qualitative Results on Recovery Step.} Figure \ref{fig_qualitative} demonstrates how our recovery step helps the VLM complete tasks despite failures. In Figure \ref{fig_qualitative} (a), the model initially generates an incorrect bounding box for an action-object interaction ($t=17$) but corrects it at $t=18$, achieving the subgoal and completing the task. In Figure \ref{fig_qualitative} (b), when a failure occurs during execution, the VLM generates a new action to complete the task. We observe that during multi-step interactions, the VLM can repeat previous actions, possibly due to token distribution flattening during SFT. The recovery step mitigates this by generating both bounding boxes and appropriate subsequent actions, improving task performance. 

We provide additional failure cases, including analyses of how removing individual skills affects performance on each task, and present {evaluation results on the validation set in LangR, a performance analysis on data scaling}, and an ablation study on the memory input and recovery step count in the supplementary material, Sec. F.
 
\vspace{-0.2cm}
\section{Conclusion}
\vspace{-0.1cm}
\label{sec:conclusion}
In this study, we propose a novel framework EUEA, which fine-tunes four core skills to VLM that enable agents enhance their environmental understanding. Our skill benchmark shows that existing VLMs still exhibit limited environmental understanding without specific environment skill fine-tuning. Additionally, we introduce a recovery step that selects an alternative action through sampling when an interaction fails, and a GRPO stage that reduces inconsistent responses. These approachs allow the model to successfully complete tasks, achieving a high success rate of 86.48\%. Our findings highlight the importance of environmental understanding, skill learning, the recovery step, and the additional refinement stage in developing more reliable end-to-end agents.

\noindent\textbf{Limitation and Future Work.} This study evaluates task performance and skill evaluation in a discrete environment, which may limit a comprehensive understanding of state transitions. Future work should extend this approach to continuous environments by incorporating video-based input, enabling better tracking of environmental changes. Additionally, while this study focuses on interaction, low-level navigation remains a crucial challenge for end-to-end agents. Moreover, current models, including ours, often rely on intuitive predictions rather than explicit reasoning, which may lead to misunderstandings of the environment. We believe that integrating reasoning mechanisms that utilize past information and predict future outcomes would enhance decision-making capabilities. 
\newpage
\noindent\section*{Acknowledgement} This work was supported by Institute of Information \& communications Technology Planning \& Evaluation (IITP) grant funded by the Korea government (MSIT) (No.RS-2025-25442824, AI Star Fellowship Program (Ulsan National Institute of Science and Technology \& No.RS-2020-II201336, Artificial Intelligence graduate school support (UNIST)) and the National Research Foundation of Korea (NRF) grant funded by the Korea government (MSIT) (No.RS-2025-24683548).

{
    \small
    \bibliographystyle{ieeenat_fullname}
    \bibliography{main}

@String(CVPR= {IEEE Conf. Comput. Vis. Pattern Recog.})

@String(ICLR = {Int. Conf. Learn. Represent.})

@String(CVPR  = {CVPR})

@String(ICLR  = {ICLR})

@article{zhao2024epo,
  title={EPO: Hierarchical LLM Agents with Environment Preference Optimization},
  author={Zhao et al.},
  journal={EMNLP},
  year={2024}
}

@article{inoue2022prompter,
  title={Prompter: Utilizing Large Language Model Prompting for a Data Efficient Embodied Instruction Following},
  author={Inoue et al.},
  journal={arXiv},
  year={2022}
}

@inproceedings{szot2023large,
  title={Large language models as generalizable policies for embodied tasks},
  author={Szot, Andrew and Schwarzer, Max and Agrawal, Harsh and Mazoure, Bogdan and Metcalf, Rin and Talbott, Walter and Mackraz, Natalie and Hjelm, R Devon and Toshev, Alexander T},
  booktitle={The Twelfth International Conference on Learning Representations},
  year={2023}
}

@inproceedings{yang2024embodied,
  title={Embodied multi-modal agent trained by an llm from a parallel textworld},
  author={Yang, Yijun and Zhou, Tianyi and Li, Kanxue and Tao, Dapeng and Li, Lusong and Shen, Li and He, Xiaodong and Jiang, Jing and Shi, Yuhui},
  booktitle={Proceedings of the IEEE/CVF conference on computer vision and pattern recognition},
  pages={26275--26285},
  year={2024}
}

@article{li2024llava,
  title={LLaVA-OneVision: Easy Visual Task Transfer},
  author={Li, Bo and Zhang, Yuanhan and Guo, Dong and Zhang, Renrui and Li, Feng and Zhang, Hao and Zhang, Kaichen and Li, Yanwei and Liu, Ziwei and Li, Chunyuan},
  journal={arXiv preprint arXiv:2408.03326},
  year={2024}
}

@article{bai2025qwen2,
  title={Qwen2. 5-VL Technical Report},
  author={Bai, Shuai and Chen, Keqin and Liu, Xuejing and Wang, Jialin and Ge, Wenbin and Song, Sibo and Dang, Kai and Wang, Peng and Wang, Shijie and Tang, Jun and others},
  journal={arXiv preprint arXiv:2502.13923},
  year={2025}
}

@inproceedings{song2023llm,
  title={Llm-planner: Few-shot grounded planning for embodied agents with large language models},
  author={Song, Chan Hee and Wu, Jiaman and Washington, Clayton and Sadler, Brian M and Chao, Wei-Lun and Su, Yu},
  booktitle={Proceedings of the IEEE/CVF international conference on computer vision},
  pages={2998--3009},
  year={2023}
}

@inproceedings{yao2023react,
  title={React: Synergizing reasoning and acting in language models},
  author={Yao, Shunyu and Zhao, Jeffrey and Yu, Dian and Du, Nan and Shafran, Izhak and Narasimhan, Karthik and Cao, Yuan},
  booktitle={International Conference on Learning Representations (ICLR)},
  year={2023}
}

@article{sun2023adaplanner,
  title={Adaplanner: Adaptive planning from feedback with language models},
  author={Sun, Haotian and Zhuang, Yuchen and Kong, Lingkai and Dai, Bo and Zhang, Chao},
  journal={Advances in neural information processing systems},
  volume={36},
  pages={58202--58245},
  year={2023}
}

@inproceedings{zhou2024embodied,
  title={Embodied understanding of driving scenarios},
  author={Zhou, Yunsong and Huang, Linyan and Bu, Qingwen and Zeng, Jia and Li, Tianyu and Qiu, Hang and Zhu, Hongzi and Guo, Minyi and Qiao, Yu and Li, Hongyang},
  booktitle={European Conference on Computer Vision},
  pages={129--148},
  year={2024},
  organization={Springer}
}

@article{shinn2023reflexion,
  title={Reflexion: Language agents with verbal reinforcement learning},
  author={Shinn, Noah and Cassano, Federico and Gopinath, Ashwin and Narasimhan, Karthik and Yao, Shunyu},
  journal={Advances in Neural Information Processing Systems},
  volume={36},
  pages={8634--8652},
  year={2023}
}

@article{li2025embodied,
  title={Embodied agent interface: Benchmarking llms for embodied decision making},
  author={Li, Manling and Zhao, Shiyu and Wang, Qineng and Wang, Kangrui and Zhou, Yu and Srivastava, Sanjana and Gokmen, Cem and Lee, Tony and Li, Erran Li and Zhang, Ruohan and others},
  journal={Advances in Neural Information Processing Systems},
  volume={37},
  pages={100428--100534},
  year={2025}
}

@article{wang2023voyager,
  title={Voyager: An open-ended embodied agent with large language models},
  author={Wang, Guanzhi and Xie, Yuqi and Jiang, Yunfan and Mandlekar, Ajay and Xiao, Chaowei and Zhu, Yuke and Fan, Linxi and Anandkumar, Anima},
  journal={arXiv preprint arXiv:2305.16291},
  year={2023}
}

@article{ahn2022can,
  title={Do as i can, not as i say: Grounding language in robotic affordances},
  author={Ahn, Michael and Brohan, Anthony and Brown, Noah and Chebotar, Yevgen and Cortes, Omar and David, Byron and Finn, Chelsea and Fu, Chuyuan and Gopalakrishnan, Keerthana and Hausman, Karol and others},
  journal={arXiv preprint arXiv:2204.01691},
  year={2022}
}

@article{levine2016end,
  title={End-to-end training of deep visuomotor policies},
  author={Levine, Sergey and Finn, Chelsea and Darrell, Trevor and Abbeel, Pieter},
  journal={Journal of Machine Learning Research},
  volume={17},
  number={39},
  pages={1--40},
  year={2016}
}

@inproceedings{shridhar2020alfred,
  title={Alfred: A benchmark for interpreting grounded instructions for everyday tasks},
  author={Shridhar, Mohit and Thomason, Jesse and Gordon, Daniel and Bisk, Yonatan and Han, Winson and Mottaghi, Roozbeh and Zettlemoyer, Luke and Fox, Dieter},
  booktitle={Proceedings of the IEEE/CVF conference on computer vision and pattern recognition},
  pages={10740--10749},
  year={2020}
}

@article{shridhar2020alfworld,
  title={Alfworld: Aligning text and embodied environments for interactive learning},
  author={Shridhar, Mohit and Yuan, Xingdi and C{\^o}t{\'e}, Marc-Alexandre and Bisk, Yonatan and Trischler, Adam and Hausknecht, Matthew},
  journal={arXiv preprint arXiv:2010.03768},
  year={2020}
}

@article{internvl2_5,
  title={Expanding performance boundaries of open-source multimodal models with model, data, and test-time scaling},
  author={Chen, Zhe and Wang, Weiyun and Cao, Yue and Liu, Yangzhou and Gao, Zhangwei and Cui, Erfei and Zhu, Jinguo and Ye, Shenglong and Tian, Hao and Liu, Zhaoyang and others},
  journal={arXiv preprint arXiv:2412.05271},
  year={2024}
}

@article{kolve2017ai2,
  title={Ai2-thor: An interactive 3d environment for visual ai},
  author={Kolve, Eric and Mottaghi, Roozbeh and Han, Winson and VanderBilt, Eli and Weihs, Luca and Herrasti, Alvaro and Deitke, Matt and Ehsani, Kiana and Gordon, Daniel and Zhu, Yuke and others},
  journal={arXiv preprint arXiv:1712.05474},
  year={2017}
}

@article{yoo2025exploratory,
  title={Exploratory Retrieval-Augmented Planning For Continual Embodied Instruction Following},
  author={Yoo, Minjong and Jang, Jinwoo and Park, Wei-Jin and Woo, Honguk},
  journal={Advances in Neural Information Processing Systems},
  volume={37},
  pages={67034--67060},
  year={2025}
}

@inproceedings{huang2022language,
  title={Language models as zero-shot planners: Extracting actionable knowledge for embodied agents},
  author={Huang, Wenlong and Abbeel, Pieter and Pathak, Deepak and Mordatch, Igor},
  booktitle={International conference on machine learning},
  pages={9118--9147},
  year={2022},
  organization={PMLR}
}

@inproceedings{choi2024lota,
  title={LoTa-Bench: Benchmarking Language-oriented Task Planners for Embodied Agents},
  author={Choi, Jae-Woo and Yoon, Youngwoo and Ong, Hyobin and Kim, Jaehong and Jang, Minsu},
  booktitle={International Conference on Learning Representations (ICLR)},
  year={2024}
}

@article{achiam2023gpt,
  title={Gpt-4 technical report},
  author={Achiam, Josh and Adler, Steven and Agarwal, Sandhini and Ahmad, Lama and Akkaya, Ilge and Aleman, Florencia Leoni and Almeida, Diogo and Altenschmidt, Janko and Altman, Sam and Anadkat, Shyamal and others},
  journal={arXiv preprint arXiv:2303.08774},
  year={2023}
}

@inproceedings{
    islam2024eqamx,
    title={{EQA}-{MX}: Embodied Question Answering using Multimodal Expression},
    author={Md Mofijul Islam and Alexi Gladstone and Riashat Islam and Tariq Iqbal},
    booktitle={The Twelfth International Conference on Learning Representations},
    year={2024},
    url={https://openreview.net/forum?id=7gUrYE50Rb}
}

@InProceedings{Majumdar_2024_CVPR,
    author    = {Majumdar, Arjun and Ajay, Anurag and Zhang, Xiaohan and Putta, Pranav and Yenamandra, Sriram and Henaff, Mikael and Silwal, Sneha and Mcvay, Paul and Maksymets, Oleksandr and Arnaud, Sergio and Yadav, Karmesh and Li, Qiyang and Newman, Ben and Sharma, Mohit and Berges, Vincent and Zhang, Shiqi and Agrawal, Pulkit and Bisk, Yonatan and Batra, Dhruv and Kalakrishnan, Mrinal and Meier, Franziska and Paxton, Chris and Sax, Alexander and Rajeswaran, Aravind},
    title     = {OpenEQA: Embodied Question Answering in the Era of Foundation Models},
    booktitle = {Proceedings of the IEEE/CVF Conference on Computer Vision and Pattern Recognition (CVPR)},
    month     = {June},
    year      = {2024},
    pages     = {16488-16498}
}

@misc{mfe-etp,
      title={MFE-ETP: A Comprehensive Evaluation Benchmark for Multi-modal Foundation Models on Embodied Task Planning}, 
      author={Min Zhang and Xian Fu and Jianye Hao and Peilong Han and Hao Zhang and Lei Shi and Hongyao Tang and Yan Zheng},
      year={2024},
      eprint={2407.05047},
      archivePrefix={arXiv},
      primaryClass={cs.AI},
      url={https://arxiv.org/abs/2407.05047}, 
}

@misc{sbert,
      title={Sentence-BERT: Sentence Embeddings using Siamese BERT-Networks}, 
      author={Nils Reimers and Iryna Gurevych},
      year={2019},
      eprint={1908.10084},
      archivePrefix={arXiv},
      primaryClass={cs.CL},
      url={https://arxiv.org/abs/1908.10084}, 
}

@article{szot2024grounding,
  title={Grounding multimodal large language models in actions},
  author={Szot, Andrew and Mazoure, Bogdan and Agrawal, Harsh and Hjelm, R Devon and Kira, Zsolt and Toshev, Alexander},
  journal={Advances in Neural Information Processing Systems},
  volume={37},
  pages={20198--20224},
  year={2024}
}

@misc{embodiedscan,
      title={EmbodiedScan: A Holistic Multi-Modal 3D Perception Suite Towards Embodied AI}, 
      author={Tai Wang and Xiaohan Mao and Chenming Zhu and Runsen Xu and Ruiyuan Lyu and Peisen Li and Xiao Chen and Wenwei Zhang and Kai Chen and Tianfan Xue and Xihui Liu and Cewu Lu and Dahua Lin and Jiangmiao Pang},
      year={2023},
      eprint={2312.16170},
      archivePrefix={arXiv},
      primaryClass={cs.CV},
      url={https://arxiv.org/abs/2312.16170}, 
}

@article{szot2021habitat,
  title={Habitat 2.0: Training home assistants to rearrange their habitat},
  author={Szot, Andrew and Clegg, Alexander and Undersander, Eric and Wijmans, Erik and Zhao, Yili and Turner, John and Maestre, Noah and Mukadam, Mustafa and Chaplot, Devendra Singh and Maksymets, Oleksandr and others},
  journal={Advances in neural information processing systems},
  volume={34},
  pages={251--266},
  year={2021}
}

@article{internvl3,
  title={InternVL3: Exploring Advanced Training and Test-Time Recipes for Open-Source Multimodal Models},
  author={Zhu, Jinguo and Wang, Weiyun and Chen, Zhe and Liu, Zhaoyang and Ye, Shenglong and Gu, Lixin and Duan, Yuchen and Tian, Hao and Su, Weijie and Shao, Jie and others},
  journal={arXiv preprint arXiv:2504.10479},
  year={2025}
}

@article{zheng2024llamafactory,
  title={Llamafactory: Unified efficient fine-tuning of 100+ language models},
  author={Zheng, Yaowei and Zhang, Richong and Zhang, Junhao and Ye, Yanhan and Luo, Zheyan and Feng, Zhangchi and Ma, Yongqiang},
  journal={arXiv preprint arXiv:2403.13372},
  year={2024}
}

@article{hu2022lora,
  title={Lora: Low-rank adaptation of large language models.},
  author={Hu, Edward J and Shen, Yelong and Wallis, Phillip and Allen-Zhu, Zeyuan and Li, Yuanzhi and Wang, Shean and Wang, Lu and Chen, Weizhu and others},
  journal={ICLR},
  volume={1},
  number={2},
  pages={3},
  year={2022}
}

@article{wang2024e2cl,
  title={E2CL: exploration-based error correction learning for embodied agents},
  author={Wang, Hanlin and Leong, Chak Tou and Wang, Jian and Li, Wenjie},
  journal={arXiv preprint arXiv:2409.03256},
  year={2024}
}

@article{duan2024aha,
  title={Aha: A vision-language-model for detecting and reasoning over failures in robotic manipulation},
  author={Duan, Jiafei and Pumacay, Wilbert and Kumar, Nishanth and Wang, Yi Ru and Tian, Shulin and Yuan, Wentao and Krishna, Ranjay and Fox, Dieter and Mandlekar, Ajay and Guo, Yijie},
  journal={arXiv preprint arXiv:2410.00371},
  year={2024}
}

@article{hu2019you,
  title={Are you looking? grounding to multiple modalities in vision-and-language navigation},
  author={Hu, Ronghang and Fried, Daniel and Rohrbach, Anna and Klein, Dan and Darrell, Trevor and Saenko, Kate},
  journal={arXiv preprint arXiv:1906.00347},
  year={2019}
}

@inproceedings{zhu2020vision,
  title={Vision-language navigation with self-supervised auxiliary reasoning tasks},
  author={Zhu, Fengda and Zhu, Yi and Chang, Xiaojun and Liang, Xiaodan},
  booktitle={Proceedings of the IEEE/CVF conference on computer vision and pattern recognition},
  pages={10012--10022},
  year={2020}
}

@article{chaplot2020object,
  title={Object goal navigation using goal-oriented semantic exploration},
  author={Chaplot, Devendra Singh and Gandhi, Dhiraj Prakashchand and Gupta, Abhinav and Salakhutdinov, Russ R},
  journal={Advances in Neural Information Processing Systems},
  volume={33},
  pages={4247--4258},
  year={2020}
}

@inproceedings{blukis2022persistent,
  title={A persistent spatial semantic representation for high-level natural language instruction execution},
  author={Blukis, Valts and Paxton, Chris and Fox, Dieter and Garg, Animesh and Artzi, Yoav},
  booktitle={Conference on Robot Learning},
  pages={706--717},
  year={2022},
  organization={PMLR}
}

@inproceedings{liu2024volumetric,
  title={Volumetric environment representation for vision-language navigation},
  author={Liu, Rui and Wang, Wenguan and Yang, Yi},
  booktitle={Proceedings of the IEEE/CVF conference on computer vision and pattern recognition},
  pages={16317--16328},
  year={2024}
}

@article{pantazopoulos2023multitask,
  title={Multitask multimodal prompted training for interactive embodied task completion},
  author={Pantazopoulos, Georgios and Nikandrou, Malvina and Parekh, Amit and Hemanthage, Bhathiya and Eshghi, Arash and Konstas, Ioannis and Rieser, Verena and Lemon, Oliver and Suglia, Alessandro},
  journal={arXiv preprint arXiv:2311.04067},
  year={2023}
}

@article{yang2025embodiedbench,
  title={Embodiedbench: Comprehensive benchmarking multi-modal large language models for vision-driven embodied agents},
  author={Yang, Rui and Chen, Hanyang and Zhang, Junyu and Zhao, Mark and Qian, Cheng and Wang, Kangrui and Wang, Qineng and Koripella, Teja Venkat and Movahedi, Marziyeh and Li, Manling and others},
  journal={arXiv preprint arXiv:2502.09560},
  year={2025}
}

@article{zhang2021hierarchical,
  title={Hierarchical task learning from language instructions with unified transformers and self-monitoring},
  author={Zhang, Yichi and Chai, Joyce},
  journal={arXiv preprint arXiv:2106.03427},
  year={2021}
}

@inproceedings{zhang2022danli,
  title={Danli: Deliberative agent for following natural language instructions},
  author={Zhang, Yichi and Yang, Jianing and Pan, Jiayi and Storks, Shane and Devraj, Nikhil and Ma, Ziqiao and Yu, Keunwoo and Bao, Yuwei and Chai, Joyce},
  booktitle={Proceedings of the 2022 Conference on Empirical Methods in Natural Language Processing},
  pages={1280--1298},
  year={2022}
}

@article{shao2024deepseekmath,
  title={Deepseekmath: Pushing the limits of mathematical reasoning in open language models},
  author={Shao, Zhihong and Wang, Peiyi and Zhu, Qihao and Xu, Runxin and Song, Junxiao and Bi, Xiao and Zhang, Haowei and Zhang, Mingchuan and Li, YK and Wu, Yang and others},
  journal={arXiv preprint arXiv:2402.03300},
  year={2024}
}

@inproceedings{szot2025multimodal,
  title={From multimodal llms to generalist embodied agents: Methods and lessons},
  author={Szot, Andrew and Mazoure, Bogdan and Attia, Omar and Timofeev, Aleksei and Agrawal, Harsh and Hjelm, Devon and Gan, Zhe and Kira, Zsolt and Toshev, Alexander},
  booktitle={Proceedings of the Computer Vision and Pattern Recognition Conference},
  pages={10644--10655},
  year={2025}
}

@article{kaelbling1998planning,
  title={Planning and acting in partially observable stochastic domains},
  author={Kaelbling, Leslie Pack and Littman, Michael L and Cassandra, Anthony R},
  journal={Artificial intelligence},
  volume={101},
  number={1-2},
  pages={99--134},
  year={1998},
  publisher={Elsevier}
}

@article{yu2025dapo,
  title={Dapo: An open-source llm reinforcement learning system at scale, 2025},
  author={Yu, Qiying and Zhang, Zheng and Zhu, Ruofei and Yuan, Yufeng and Zuo, Xiaochen and Yue, Yu and Fan, Tiantian and Liu, Gaohong and Liu, Lingjun and Liu, Xin and others},
  journal={URL https://arxiv. org/abs/2503.14476},
  year={2025}
}

@article{feng2025group,
  title={Group-in-group policy optimization for llm agent training},
  author={Feng, Lang and Xue, Zhenghai and Liu, Tingcong and An, Bo},
  journal={arXiv preprint arXiv:2505.10978},
  year={2025}
}

@article{chen2025training,
  title={Training Strategies for Efficient Embodied Reasoning},
  author={Chen, William and Belkhale, Suneel and Mirchandani, Suvir and Mees, Oier and Driess, Danny and Pertsch, Karl and Levine, Sergey},
  journal={arXiv preprint arXiv:2505.08243},
  year={2025}
}

@misc{intelligence2025pi05visionlanguageactionmodelopenworld,
      title={$\pi_{0.5}$: a Vision-Language-Action Model with Open-World Generalization}, 
      author={Physical Intelligence and Kevin Black and Noah Brown and James Darpinian and Karan Dhabalia and Danny Driess and Adnan Esmail and Michael Equi and Chelsea Finn and Niccolo Fusai and Manuel Y. Galliker and Dibya Ghosh and Lachy Groom and Karol Hausman and Brian Ichter and Szymon Jakubczak and Tim Jones and Liyiming Ke and Devin LeBlanc and Sergey Levine and Adrian Li-Bell and Mohith Mothukuri and Suraj Nair and Karl Pertsch and Allen Z. Ren and Lucy Xiaoyang Shi and Laura Smith and Jost Tobias Springenberg and Kyle Stachowicz and James Tanner and Quan Vuong and Homer Walke and Anna Walling and Haohuan Wang and Lili Yu and Ury Zhilinsky},
      year={2025},
      eprint={2504.16054},
      archivePrefix={arXiv},
      primaryClass={cs.LG},
      url={https://arxiv.org/abs/2504.16054}, 
}

@article{liang2022code,
  title={Code as policies: Language model programs for embodied control},
  author={Liang, Jacky and Huang, Wenlong and Xia, Fei and Xu, Peng and Hausman, Karol and Ichter, Brian and Florence, Pete and Zeng, Andy},
  journal={arXiv preprint arXiv:2209.07753},
  year={2022}
}

@article{bai2025qwen3,
  title={Qwen3-vl technical report},
  author={Bai, Shuai and Cai, Yuxuan and Chen, Ruizhe and Chen, Keqin and Chen, Xionghui and Cheng, Zesen and Deng, Lianghao and Ding, Wei and Gao, Chang and Ge, Chunjiang and others},
  journal={arXiv preprint arXiv:2511.21631},
  year={2025}
}
}

\end{document}